\documentclass[11pt]{article}

\usepackage[utf8]{inputenc} 
\usepackage{hyperref}       
\usepackage{url}            
\usepackage{booktabs}       
\usepackage{amsfonts}       
\usepackage{nicefrac}       
\usepackage{microtype}      
\usepackage{lipsum}
\usepackage{cite}
\usepackage{amsmath, amsthm, amssymb}
\usepackage{mathtools}
\usepackage[margin=2.9cm]{geometry}
\usepackage{authblk}
\usepackage{threeparttable}
\usepackage{graphicx}
\usepackage{float}

\title{An Exploratory Analysis of Biased Learners in Soft-Sensing Frames}

\author[1]{Aysun Urhan\thanks{aysun.urhan@boun.edu.tr}}
\author[1]{Burak Alakent\thanks{burak.alakent@boun.edu.tr (Corresponding author)}}
\affil[1]{Department of Chemical Engineering, Bogazici University}

\begin{document}
\maketitle

\begin{abstract}
Data driven soft sensor design has recently gained immense popularity, due to advances in sensory devices, and a growing interest in data mining. While partial least squares (PLS) is traditionally used in the process literature for designing soft sensors, the statistical literature has focused on sparse learners, such as Lasso and relevance vector machine (RVM), to solve the high dimensional data problem. In the current study, predictive performances of three regression techniques, PLS, Lasso and RVM were assessed and compared under various offline and online soft sensing scenarios applied on datasets from five real industrial plants, and a simulated process. In offline learning, predictions of RVM and Lasso were found to be superior to those of PLS when a large number of time-lagged predictors were used. Online prediction results gave a slightly more complicated picture. It was found that the minimum prediction error achieved by PLS under moving window (MW), or just-in-time learning scheme was decreased up to $\sim$5-10\% using Lasso, or RVM. However, when a small MW size was used, or the optimum number of PLS components was as low as $\sim$1, prediction performance of PLS surpassed RVM, which was found to yield occasional unstable predictions. PLS and Lasso models constructed via online parameter tuning generally did not yield better predictions compared to those constructed via offline tuning. We present evidence to suggest that retaining a large portion of the available process measurement data in the predictor matrix, instead of preselecting variables, would be more advantageous for sparse learners in increasing prediction accuracy. As a result, Lasso is recommended as a better substitute for PLS in soft sensors; while performance of RVM should be validated before online application.
\end{abstract}


\section{Introduction}
In process industries, data driven soft sensors can provide real-time predictions for process control, monitoring, validating hardware measurements \cite{1, 2, 3, 4, 5, 6}. Since predictive performance of soft sensors built in offline mode deteriorates in time, due to changes in raw materials, catalyst deactivation, fouling, and seasonal effects \cite{7}, referred to as “concept drift”, online learning schemes are used in soft sensor design to maintain the desired predictive accuracy \cite{8}. In machine learning applications, various adaptive learning frames, such as recursive methods, moving window (MW), just-in-time learning (JITL) and ensemble modeling, have been suggested for model adaptation and online prediction \cite{9,10,11,12,13}. Another challenge in process industries is the large number of correlated measured variables, since increasing the number of predictors leads to an inflation in variance of parameter estimates. Hence, designing an accurate soft sensor lies not only in devising a convenient adaptive scheme, but also using a learner tailored for the specifics of the process.\\

Principle component regression (PCR) and partial least squares (PLS) have been the traditional latent variable learners used in statistical modeling of steady state and dynamic processes\cite{14,15,16,17}. PLS has been historically expected to yield models with high stability, obtained through “maximal reduction in the covariance of the data”\cite{18}. More recent machine learning tools, such as least squares support vector machines (SVMs), Gaussian process regression, Bayesian networks and extreme learning machine are being successfully incorporated into soft-sensor design in chemical processes, particularly to cope with the nonlinearity issue \cite{27,73,9}. However, most of the adaptive learning algorithms in the literature have relied on traditional PLS/PCR for prediction, mainly due to the following advantages of linear (in parameters) models: (i) ease of interpretation of the prediction results \cite{new1,new2,new3,new4} (ii) local linear learners are prevalent for implementation in inferential/predictive controllers \cite{new5,new6,new7,new8} and (iii) frequency of concept drifts in chemical processes entails using a small, relevant subset of the data, in which linear models yield better predictive models \cite{19,20,21,22,23,24,26,28}. One of the first serious criticisms to PLS was vocalized by Frank and Friedman; they suggested that prediction quality of PLS is likely to deteriorate when a large number of components are included, due to shrinkage factors being larger than unity along the eigendirections, i.e. “expanding” the LS estimates, and advocated using ridge regression (RR)\cite{29}. To this criticism, it was stated that shrinkage factors of PLS is within [0, 1] interval in the signal-to-noise ratio directions, i.e. directions which take the LS point estimates and the variance into consideration\cite{30,31}, giving a theoretical basis for the predictive performance of PLS. Hastie and coworkers continued with the criticism, suggesting that covariance of the predictors becomes more important than correlation with the response variables in determining latent variables via PLS method\cite{32}. Consistent with this view, mathematical analyses and simulations showed that predictions of PLS deteriorate for larger number of predictors, hence subset selection prior to using PLS was recommended\cite{33,34,35}. \\

To tackle the multicollinearity problem, statistics community has been developing biased estimators (or shrinkage/regularization methods). The oldest of these methods is RR, in which the least squares (LS) estimation is modified by imposing an L2 penalty onto the regression coefficients\cite{36}. It should be noted that PCR and PLS are also shrinkage methods, which affect the contribution of eigendirections of the predictor and predictor-response variable covariance matrices, respectively, to the parameter estimates. Least absolute shrinkage and selection operator (Lasso) works in a similar way, the only difference being that L2 penalty in RR is replaced with an L1 penalty, which may lead to zero coefficient estimates and yield sparse models\cite{37}. Similarly, relevance vector machine (RVM) was first introduced as a sparse Bayesian model for both regression and classification tasks; motivated by Bayesian automatic relevance determination, RVM uses the same functional form as SVM\cite{38}. Sparse PCA and PLS were more recently developed to fuse latent vector representation with predictor selection via L1/L2 penalties\cite{34,39}. In high dimensional problems ranging from computer vision to chemometrics, as well as gene expression, sparse models have been noted for improved generalization capability\cite{40,41,42,43}. Recently, sparse regression techniques have been employed in process literature\cite{44,45,46,47}. However, most of these applications focus on feature selection; e.g. Lasso has been analyzed as an alternative to conventional feature selection methods for PLS based soft sensor models\cite{48,49,50}. Similarly, prediction potential of RVM, along with its ability to estimate uncertainty in predictions\cite{51}, is yet to be exploited in applications to processes.\\

Various adaptive learning methods have already been devised successfully for industrial processes, however, to our knowledge, there is not a sufficient number of studies, in which predictive performances of biased learners have been compared under identical learning settings and process environment. Additionally, studies in statistical literature, comparing the prediction accuracies of various learners, are usually based on Monte Carlo simulations, benchmark data from machine learning repository or biological data, hence may not adequately represent process engineering environment\cite{29,34,37,38,39}. As a result, the most suitable learner for soft sensor design in process industries still remains an open question. We tackled this problem in the current study via comparing the predictive performances of PLS, Lasso and RVM under offline/online learning schemes on five publicly available datasets of real chemical processes, and a continuous stirred tank (CSTR) simulation. We specifically focused on PLS, Lasso and RVM due to interpretability of the resulting models, since all these three learners express the output as a linear combination of predictors. We found it more appropriate to make learner comparisons on simpler learning “modules”, such as MW and JITL schemes, compared to specific and advanced learning schemes, e.g. state-of-the art ensemble models\cite{9, 52, 53}, for the obtained results to have higher generalizability. Additionally, we worked on a number of questions that particularly pose challenge to industrial practitioners, such as comparison of predictive performances of offline/online learning methods using time lagged process variables, or comparison of offline/online parameter tuning. Details of the learning schemes and information on datasets and model assessment are discussed in Sections 2 and 3, respectively, while predictive performances of learners in offline/online applications are presented in Section 4. Based on a diverse sample of industrial and simulated data, our results indicate that preference for PLS based predictive models in the literature is justified under a narrow range of conditions, while Lasso and, with some reservations, RVM are advantageous in a broader parameter range, particularly, in the MW frame. 

\section{Background}
\label{sec:background}
Consider a dataset $D$ of historical plant operations; for a total of $N$ observations in time order, $D = \left\{ \textbf{x}_t,\textbf{y}_t \right\}_{t=1}^N$ and $\textbf{x}_t \in \mathbb{R}^p$,  $\textbf{y}_t \in \mathbb{R}^m$, where $p$ is the number of predictors, and $m$ is the number of response variables. For a single response variable ($m = 1$), the predictor (design) matrix, \textbf{X}, is of size $N \times p$, and the response vector \textbf{y} is $N \times 1$. Using the subscript $\tau$, a subset of $D$, comprising successive observations with time indices defined in $\tau$, is represented as $D_{\tau} = \left\{ \textbf{x}_t,\textbf{y}_t \right\} $, $t \in \tau$, satisfying $\left| \tau \right| < N $. In a supervised learning setting, it is presumed that labeled outputs, $\textbf{y}$, are a function of the inputs, $\textbf{X}$, hence $\hat{\mathbf{y}}_t = \hat{f} \left( \textbf{x}_t, t \right)$ forms the basis of regression. In the current study, a single observation of $r$ process variables sampled at time $t$ is shown by $\text{z}_i (t)$, $i = 1, 2, ..., r$. Hence, the finite impulse response (FIR) parametrization of $\hat{f}$ is constructed as $\textbf{x}^{\text{T}}_t = \left[ \text{z}_1(t), \hdots , \text{z}_r(t), \text{z}_1(t-1), \hdots , \text{z}_r(t-1), \text{z}_1(t-n), \hdots , \text{z}_r(t-n) \right]$ with $p$ elements equal to $(n+1)\times r$, and the response variable is predicted using:
\begin{align}
\hat{\mathbf{y}}_t = \mathbf{x}^{\text{T}}_t \hat{\mathbf{\beta}}
\label{eq:1}
\end{align}
 
\subsection{Statistical Learning Methods}

Statistical learning methods (or learners) can be used to estimate the model parameters in Eq. \ref{eq:1} from training data. One way to decrease the mean squared error (MSE) of predictions is to use regularized regression methods\cite{29}. The current study focuses on three of these regression methods: PLS, Lasso, and RVM.

\subsubsection{Partial Least Squares (PLS)}

PLS was first introduced by Wold\cite{54}, and has been among the most popular modeling methods in soft sensor design. PLS extracts orthogonal latent score vectors, $\textbf{t}_l$ (for $l = 1, 2, \hdots , L$ where $L$) is the dimension of the latent space in predictor space, with high variance and high correlation with the response vector. PLS components are usually determined via the NIPALS algorithm, initialized with auto-scaled training data, $\textbf{X}$ and $\textbf{y}$, iteratively projecting $\textbf{y}$ residuals onto $\textbf{X}$ residuals, and estimating a linear inner relation \cite{55}. In a Bayesian frame, PLS shrinks more along the higher indexed principal directions \cite{29}. A more informative formulation for PLS is given by the following criterion, in which the $l^{\text{th}}$ weight is computed as\cite{56}:

\begin{align}
\mathbf{c}_{l}&=\underset{\mathbf{c}}{\operatorname{argmax}}\left(\operatorname{corr}\left(\mathbf{y}, \mathbf{c}^{\text{T}} \mathbf{X}\right)^{2} \operatorname{var}\left(\mathbf{c}^{\text{T}} \mathbf{X}\right)\right) \label{eq:2} \\
& \text { s.t. } \mathbf{c}^{\text{T}} \mathbf{c}=1, \, \mathbf{c}^{\text{T}} \mathbf{X}^{\text{T}} \mathbf{X} \mathbf{c}_{j}=0 \quad j = 1, 2, \ldots l-1 \, \nonumber
\end{align}

Here, $\textbf{t}_l$ score can be determined via projecting $\textbf{X}$ on $\mathbf{c}_l$. The value of $L$ is usually determined via cross validation (CV) \cite{57}.

\subsubsection{Least Absolute Shrinkage and Selection Operator (Lasso)}

Regularized regression methods use a penalty term in the objective function to prevent inflation of coefficient estimates; RR \cite{36} and Lasso \cite{37} use L2 ($q = 2$) and L1 ($q = 1$) norms, respectively:
\begin{align}
\widehat{\boldsymbol{\beta}}=\underset{\beta}{\operatorname{argmin}}\left\{\frac{1}{2} \sum_{i=1}^{N}\left(\mathrm{y}_{i}-\mathbf{x}_{i} \boldsymbol{\beta}\right)^{2}+\lambda \sum_{j=1}^{p}\left|\beta_{j}\right|^{q}\right\}
\label{eq:3}
\end{align}

Here, $\lambda$ represents the extent of regularization, and can be determined via CV. From a Bayesian view, Lasso gives a Laplacian prior to the parameter estimates. Shrinkage of RR along the higher indexed eigendirections in the predictor space yields balanced and simultaneous contributions of correlated predictors to the regression model, while L1 norm in Lasso tends to select single predictors among highly correlated ones. To remedy the lack of “grouping effect” in Lasso, elastic net method, using both L1 and L2 penalties, was proposed \cite{58,59}; however, elastic net requires two parameters to be tuned, making it difficult to be applicable for online learning applications. Instead of solving the optimization problem in Eq. \ref{eq:3}, Lasso estimates are obtained using the least angle regression (LARS) path \cite{60}, or gradient descent algorithm \cite{61}. LARS algorithm further made it possible to give an alternative interpretation to Lasso solution path; LARS allows for the contribution of predictors in a stagewise manner, i.e. incrementally, as opposed to forward stepwise selection methods, in which a predictor is totally included at each step \cite{62}.

\subsubsection{Relevance Vector Machine (RVM)}

RVM is based on the linear form of regression problem for the response (target) variable y, it can be formulated as $\mathrm{y}=g(\mathbf{x} ; \mathbf{w})+\epsilon $, where $\epsilon$ is the independent zero-mean Gaussian distributed random noise with variance $\sigma^2$, hence $ \epsilon \sim \mathcal{N}\left(0, \sigma^{2}\right)$ $\mathbf{x}$ denotes the predictor values as previously shown, while $\mathbf{w}$ is a random vector of weights, and $g$ is the generative function underlying the input-output relationship \cite{38}. Consequently, $\mathrm{y} | \mathbf{x}, \mathbf{w} \sim \mathcal{N}\left(\mathrm{y} | g(\mathbf{x}, \mathbf{w}), \sigma^{2}\right)$, and the probability distribution of the response vector $\mathrm{y}$ comprised of $N$ observations is as follows:

\begin{align}
\mathrm{p}\left(\mathbf{y} | \mathbf{X}, \mathbf{w}, \sigma^{2}\right)=\left(2 \pi \sigma^{2}\right)^{(-N / 2)} \exp \left\{-\frac{1}{2 \sigma^{2}}\|\mathbf{y}-\mathbf{\Phi} \mathbf{w}\|^{2}\right\}
\label{eq:4}
\end{align}

Here, $\mathbf{w}=\left[w_{1}, w_{2}, \ldots, w_{N}\right]^{\mathrm{T}}$ and $ \mathbf{\Phi}$ is the $N \times (N+1)$ design matrix with columns $\phi\left(\mathbf{x}_{n}\right)= \left[1, K\left(\mathrm{x}_{n}, \mathrm{x}_{1}\right), K\left(\mathrm{x}_{n}, \mathrm{x}_{2}\right) \ldots, K\left(\mathrm{x}_{n}, \mathrm{x}_{N}\right)\right]^{\mathrm{T}}$ formed by the basis functions obtained from kernel $K$. Omitting conditioning on $\mathbf{x}$ to simplify the notation, RVM defines prior distribution over $\mathbf{w}$ as follows:
\begin{align}
\mathrm{p}(\mathbf{w} | \boldsymbol{\alpha})=\left(2 \pi \sigma^{2}\right)^{(-p / 2)} \Pi_{i=1}^{p} \alpha_{i}^{1 / 2} \exp \left\{-\frac{\alpha_{i} \alpha_{i}^{2}}{2 \sigma^{2}}\right\}
\label{eq:5}
\end{align}
Here $\boldsymbol{\alpha}$ is a hyperparameter vector with $p+1$ elements, each of which corresponds to the elements in $\mathbf{w}$. Posterior distribution of the weights is obtained as:
\begin{align}
\mathrm{p}\left(\mathbf{w} | \mathbf{y}, \boldsymbol{\alpha}, \sigma^{2}\right)&=\frac{\mathrm{p}\left(\mathbf{y} | \mathbf{w}, \sigma^{2}\right) \mathbf{p}(\mathbf{w} | \alpha)}{\mathrm{p}\left(\mathbf{y} | \alpha, \sigma^{2}\right)} \nonumber \\
&=(2 \pi)^{-(N+1) / 2}|\mathbf{\Sigma}|^{-1 / 2} \exp \left\{-\frac{1}{2}(\mathbf{w}-\boldsymbol{\mu})^{\mathrm{T}} \mathbf{\Sigma}^{-1}(\mathbf{w}-\boldsymbol{\mu})\right\}
\label{eq:6}
\end{align}

Here, $\mathbf{\Sigma}=\left(\sigma^{-2} \mathbf{\Phi}^{\mathrm{T}} \mathbf{\Phi}+\mathbf{A}\right)^{-1}, \boldsymbol{\mu}=\sigma^{-2} \mathbf{\Sigma} \mathbf{\Phi}^{\mathrm{T}} \mathbf{y}$ and $\mathbf{A}=\operatorname{diag}\left(\alpha_{1}, \alpha_{2}, \ldots \alpha_{N}\right) .$ Let $\mathrm{y}_{0}$ denote the response value of a test point, and inference is made via Bayes' rule.

\begin{align}
\mathrm{p}\left(\mathrm{y}_{0} | \alpha, \sigma^{2}\right) &=\int \mathrm{p}\left(\mathrm{y}_{0} | \mathbf{w}, \sigma^{2}\right) \mathrm{p}(\mathbf{w} | \boldsymbol{\alpha}) d \mathbf{w} \nonumber \\
& =(2 \pi)^{-N / 2}\left|\sigma^{2} \mathbf{I}+\mathbf{\Phi} \mathbf{A}^{-1} \mathbf{\Phi}^{\mathrm{T}}\right|^{-1 / 2} \exp \left\{-\frac{1}{2} \mathbf{y}^{\mathrm{T}}\left(\sigma^{2} \mathbf{I}+\mathbf{\Phi} \mathbf{A}^{-1} \mathbf{\Phi}^{\mathrm{T}}\right)^{-1} \mathbf{y}\right\} 
\label{eq:7}
\end{align}

Eq. \ref{eq:6} is maximized iteratively w.r.t. parameters $\boldsymbol{\alpha}$  and $\sigma^2$; $\boldsymbol{\alpha} = \boldsymbol{\alpha}_{MP}$. As most elements of $\boldsymbol{\alpha}$ go to infinity, the corresponding weights in $\mathbf{w}$ are also concentrated at 0, hence a sparse $\mathbf{w}$ estimate, with only the relevant basis functions can be obtained \cite{38, 63}. Since RVM yields sparse solutions, not only it generalizes fairly well on unseen data, but it also helps combat collinearity among predictor variables by retaining only the relevant ones. Moreover, RVM does not require any additional step for model selection or parameter tuning, and provides probabilistic predictions instead of point estimates; the error variance estimated during training can be used to assess the uncertainty in predictions.

\subsection{Offline vs Online Learning}

Offline (or batch) learning is employed using a static training set; i.e. a single predictive model is estimated using all training observations available in the training set, and used for predicting all test observations. In the presence of concept drifts, a straightforward partitioning of the historical data into successive training and test sets in offline learning yields gross prediction errors. To remedy this and obtain meaningful predictions in the current study, datasets were divided into a number of successive regions (segments), in which approximately constant concepts may be assumed to be governing the processes during a single time segment. In each of these segments, data were divided into two contiguous parts as training/test data. 

In an online learning scenario, test observations (also referred to as query points) are received sequentially. Query points are predicted in real-time and then appended to the training set (TS) when the corresponding label information becomes available \cite{64}. In the current study, evaluation on a rolling forecasting origin (one-step-ahead prediction) was performed \cite{65}; as soon as the response variable is predicted, the query point including the measured value of the response variable was appended to the training set. In the soft sensor design literature, recursive partial least squares (RPLS), MW techniques, JITL, temporal difference learning and ensemble modeling have been among the most popular adaptive learning methods \cite{14,20,28,66,67}. In the current study, we focused on MW and JITL methods, due to their simplicity, popularity and frequent use in ensemble modeling. 

In MW modeling frame, a model $\hat{f}_{\text{MW}}$, is estimated using the most recent observations $D_{\text{MW}}$ to predict the query point received at time $k$:
\begin{align}
D_{\mathrm{MW}} \equiv\left\{\left(\mathbf{x}_{t}, \mathrm{y}_{t}\right)\right\}_{t=k-\mathrm{W}+1}^{k-1}
\label{eq:8}
\end{align}

In the current study, performance of MW models with respect to different W values was elucidated. As discussed in the following sections, MW modeling with constant windows was found to be ineffective against the “harsh” drifts in the debutanizer column dataset, so an adaptive MW method, recently suggested within a novel MW-JITL learning frame in our currently unpublished work, was utilized in the current study. To employ this method, Mahalanobis distance of the query point ($d_{\text{M}}^2$), defined as follows, is used to determine whether the query point is close to the operating point defined by samples in the $D_{\text{MW}}$.

\begin{align}
d_{\mathrm{M}}^{2}\left(\mathbf{x}_{\mathrm{MW}}\right)=\left(\mathbf{x}_{k}-\overline{\mathbf{x}}_{\mathrm{MW}}\right) \mathbf{S}_{\mathrm{MW}}^{-1}\left(\mathbf{x}_{k}-\overline{\mathbf{X}}_{\mathrm{MW}}\right)
\label{eq:9}
\end{align}

Here, $\mathbf{X}_{\mathrm{MW}}$ denotes the predictor matrix of observations in $D_{\mathrm{MW}}$ in Eq. \ref{eq:8} at time $k,$ and $\overline{\mathbf{X}}_{\mathrm{MW}}$ and $\boldsymbol{S}_{\mathrm{MW}}$ denote the mean vector and covariance matrix estimated using $D_{\mathrm{MW}}$ and $\mathbf{x}_{k}$ is the query point. For normal data, upper bound of $d_{\mathrm{M}}^{2}$ at a significance level $\alpha$ is computed as follows \cite{68}:

\begin{align}
\mathrm{UB}_{\alpha}=\frac{p(\mathrm{W}-1)(\mathrm{W}+1)}{\mathrm{W}(\mathrm{W}-p)} \mathcal{F}_{p, \mathrm{W}-p, \alpha}
\label{eq:10}
\end{align}

A significantly large distance value (a positive test result at significance $\alpha=0.01$) suggests that the query point does not belong to the distribution defined by $D_{\mathrm{MW}}$, hence the current window of observations may not be suitable to estimate a predictive model. If a positive test result is encountered, the current window size is enlarged by 20\% with the addition of past observations, until the Mahalanobis distance of the query point falls below the threshold (see Algorithm S1 Supplementary Material).

In JITL, a local model is estimated using observations most similar in the predictor space selected from the entire historical database\cite{69,70}. Once, a new model is estimated for each query point, it is discarded after prediction\cite{28,71}. The most common metric used to measure dissimilarity is Euclidean distance ($d_{\mathrm{E}}^{2}$) between the query point $\mathbf{x}_k$ and any point $\mathbf{x}_i$ in the historical set: 
\begin{align}
d_{\mathrm{E}}^{2}\left(\mathbf{x}_{k}, \mathbf{x}_{i}\right)=\left(\mathbf{x}_{k}-\mathbf{x}_{i}\right)^{\mathrm{T}}\left(\mathbf{x}_{k}-\mathbf{x}_{i}\right)
\label{eq:11}
\end{align}

The number of nearest neighbors (NN), denoted with $\vert \mathrm{NN} \vert$, around the query point is set. In the current study, JITL models were constructed within convenient ranges of $\vert \mathrm{NN} \vert$ values.
\section{Datasets, Learning Scenarios, and Assessment of Learners}

In the current study, five real industrial datasets (Datasets 1 to 5) and a simulated dataset (Dataset 6) were used in comparing the predictive capabilities of the learners. A summary of the industrial datasets is given in Table 1, in which $N, r, m$ and maximum VIF denote the total number of samples, number of available process (input) variables, the number of quality variables to be predicted, and the maximum variance inflation factors of the predictor matrix constructed from the process variables, respectively, while details of all the processes are discussed in Section 3.1. It should be noted that VIF values reported here are much less than those “experienced” by the learners during the adaptive learning procedure (see Results and Discussion). Details of the assessment of the learners under offline and online learning schemes are discussed in Section 3.2. 

\begin{table}
   \caption{A summary of datasets used in the experiments.}
   \vskip\baselineskip
   \label{table:1}
   \begin{threeparttable}[b]
	\renewcommand{\TPTminimum}{\linewidth}
    \makebox[\linewidth]{%
	\begin{tabular}{lllll}
    \toprule
    Name & $N$ & $r$ & $m$ & Maximum VIF\tnote{1} \\
    \midrule
	DS1  & 10,081 & 5 & 2 & 10.1 \\
	DS2	 & 2,394 & 7 & 1 & 61.2 \\
	DS3  & 360 & 11 & 1 & 17.5 \\
	DS4\tnote{2} & 331 & 12/8/9 & 3 & 15.5/17.6/19.5 \\
	DS5	 & 104 & 6 & 3 & 6.17 \\
	DS6/DS6p\tnote{3} & 700 (8$\times$20)\tnote{4} & 19/10 & 1 & 1.0$\times$10\textsuperscript{4}/626 \\
    \bottomrule
  \end{tabular}}    
  \begin{tablenotes}
    \item[1]{\small VIF values are computed for the whole 		datasets.}
	\item[2]{\small Parameters and VIF values for each of the three response variables are separated by slash.}
	\item[3]{\small Parameters and VIF values of DS6 and preprocessed DS6 (DS6p) are separated by a slash.}
	\item[4]{\small DS6/DS6p comprise 8 different CDMs, each repeated for 20 times and contain 700 time-ordered samples.}
  \end{tablenotes}
  \end{threeparttable}
\end{table}

\subsection{Industrial and Simulated Datasets}
\paragraph{Dataset 1 (DS1): Sulfur recovery unit}
A sulfur recovery unit with two combustion chambers is used to remove polluting agents from acid gas, while elemental sulfur is produced as a by-product\cite{3}. For process monitoring, soft sensors are designed to predict the concentrations of H2S and SO2, which, otherwise, may severely damage the sensors. Historical data comprises 10,081 observations of five process variables, consisting of gas flowrates in four ammonia rich inlet streams, and air flowrate in a supplementary air stream, and two response variables sampled at 1 min intervals.

\paragraph{Dataset 2 (DS2): Debutanizer column}
The tail stream of the sulfur recovery plant, from which DS1 was obtained, is processed to separate ethane and butane in a sequence of distillation columns\cite{3}. The employed measurement scheme on butane concentration in bottoms product of the debutanizer column leads to a delay of $\sim$45 mins, so a soft sensor is designed to predict butane concentration online. Historical data consists of 2394 measurements of seven input variables, which include tray/bottom temperatures, reflux/distillate flowrates and column pressure, sampled at $\sim$6 min intervals.

\paragraph{Dataset 3 (DS3): Waste water treatment plant}
A real-world urban water treatment plant consists of pre-treatment, coagulation, flocculation, sedimentation, filtration and disinfection processes\cite{72}. Chlorine is added to waste water to remove metals, and water is clarified by removing the suspended particles. The treatment process also decreases the fluoride concentration in water; hence it is desired to make up for the fluoride that has been lost. A soft sensor is to be designed to predict fluorine concentration, which is measured via laboratory analyses only at every 24 hours. Historical dataset consists of 360 observations of 11 input variables, e.g. physical properties of water streams, such as turbidity and color, and pH level, and their most recent, two and four time lagged measurements, sampled at every two hours.
 
\paragraph{Dataset 4 (DS4): Melt index}
A sequential-reactor-multi-grade polyethylene production process is used in an industrial polymerization plant located in Taiwan\cite{25}. The polymer manufacturing process is suggested to be challenging due to its sequential setup, and frequent changes in the operating conditions since different grades of polymer are produced. Soft sensors are to be used to predict melt indices of products from three reactors that can be measured only once a day via offline laboratory analyses. There are 331 observations in the historical data, consisting of 12, 8 and 9 process variables (details of the measured variables were not given in the original study due to confidentiality) from each reactor in that order, and one response variable for each reactor.

\paragraph{Dataset 5 (DS5): Industrial fluidized catalytic cracking unit}
Fluidized catalytic cracking is used to convert heavy gas oils to lighter hydrocarbon products with high value, hence it is among the most profitable units in a refinery. In a refinery in China,  measurement of three key components, gasoline, light diesel oil and LPG, takes 8 hours to one day, therefore soft-sensors are required to make online predictions of these variables\cite{73}. The historical dataset contains 104 observations, six process variables, which include flowrate and temperature measurements, and three response variables.

\begin{figure}[H]
\label{fig:1}
\caption{Trajectories of various measured process variables in DS6.}
\vskip\baselineskip
\centering
\includegraphics[scale=.8]{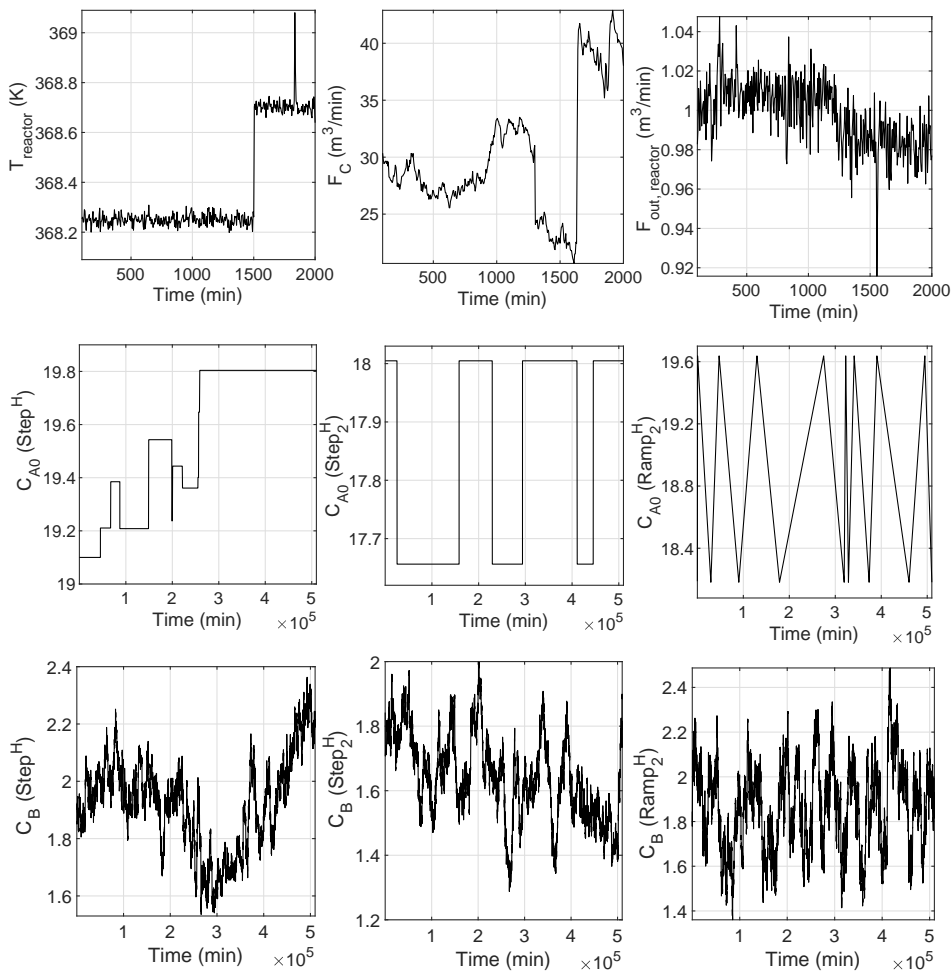}
\end{figure}

\paragraph{Dataset 6 (DS6): Synthetic data from a heat exchanger + CSTR system}
An isothermal CSTR process from the literature\cite{74} was expanded to include an external heat exchanger unit, in which the solvent feed is preheated before it enters the reactor (Fig. S1). An exothermic first order reaction, A $\rightarrow$ B, takes place inside the reactor, while the temperature and height of the reactor are controlled via manipulating the flowrate of coolant and the reactor outlet, respectively (Fig. \ref{fig:1}). Model equations (Text S1 and Tables S1-S4) were solved using MATLAB \textsuperscript{TM} Simulink for nine different simulation scenarios: eight of which represent different types of real concept drifts often encountered in chemical processes, and one simulation model with no real concept drift (Table S5). Trajectories of the unmeasured CA\textsubscript{0} values in Fig. \ref{fig:1} are representative examples of the concept drift scenarios. The aim is to design a soft sensor to predict the outlet concentration of product B using 19 measured process variables. Each simulation comprised 700 observations, which were split in 300/400 to obtain training/test sets, and the simulations were repeated 20 times for all nine CD models (CDMs). To observe the effect of preprocessing the data via selecting a subset of predictors, a second dataset (DS6p) was formed using 10 of the process variables which are most correlated with the response variable. As a result, the maximum VIF of the process variables in the DS6p was found to be $\sim$20 times smaller than that in DS6 (Table \ref{table:1}).

\subsection{Assessment of the predictive performance of learners}

Three learners, PLS, Lasso and RVM, were used to predict each quality variable separately in both offline and online learning procedures. In offline learning, predictive performance of the learners were tested using high dimensional FIR models. Number of predictors in FIR models, and sizes of training/test sets were the important meta-parameters in offline learning stage. In DS1 and DS2, input matrix were formed from process variables sampled up to $n = 24$ and 32 time lags, respectively, i.e. FIR models for DS1 and DS2 were constructed using $p = 5, 10, \ldots \, 115$ and $7, 14, \ldots \, 119$ predictors, respectively (Table \ref{table:2}). Two different training set sizes $N_{\text{TS}}$ were used in DS1 and DS2; the smaller one consisted of 250 observations, so that stable models would be obtained using $>110$ predictors. Training sets were extended to include 450 observations to see the how the performance of learners change with respect to training set size. In DS3, DS4 and DS5, only the current samples were used in the FIR input matrix, while FIR input matrix contained up to 8 lagged terms in DS6. In DS1, DS2 and DS3, datasets were divided into 20, 4 and 3 successive regions, respectively, while no segmentation of data was performed in DS4 and DS5 due to insufficient number of data points. In all the industrial datasets, small test sets ($\sim$50) were used for offline learning to be effective in the presence of concept drifts, while repetitions using different segments is expected to make up for the high variance of prediction accuracies of small test sets. In simulated dataset DS6, offline learning was employed to data sampled in the absence of real concept drift. 

\begin{table}
   \caption{Meta-parameters of the offline learning scenario.}
   \vskip.5\baselineskip
   \label{table:2}
   \begin{threeparttable}[b]
	\renewcommand{\TPTminimum}{\linewidth}
    \makebox[\linewidth]{%
	\begin{tabular}{lllll}
    \toprule
    Name & $n$ & $p$ & Number of segments & Training-test sizes \\
    \midrule
	DS1  & $\{ 0, 1, \ldots \, 24 \}$ & $\{5, 10, \ldots \, 115 \}$ & 20 & 250/450-50 \\
	DS2	 & $\{ 0, 1, \ldots \, 32 \}$ & $\{7, 14, \ldots \, 217 \}$ & 4 & 250/450-50 \\
	DS3  & 0 & 11 & 3 & 70-50 \\
	DS4  & 0 & 12/20/29\tnote{1} & 3 & 80-30 \\
	DS5	 & 0 & 6 & 1 & 40-64 \\
	DS6/DS6p & $\{ 0, 1, \ldots \, 8 \}/0$\tnote{2} & $\{ 19, 38, \ldots \, 152 \}/10$ & 1 & 300-400 \\
    \bottomrule
  \end{tabular}}    
  \begin{tablenotes}
    \item[1]{\small Each predictor set was used for one of the response variables.}
	\item[2]{\small Predictor matrix in DS6p consisted of 10 process variables sampled at zero time lag.}
  \end{tablenotes}
  \end{threeparttable}
\end{table}

For online learning assessment of the learners, which were denoted using the subscript MW or JITL, we focused on two meta-parameters: the size of the predictor matrix for the FIR model, the size of initial TS. The former was handled in a lower resolution in online learning compared to offline learning, since the main focus in online modeling is to maintain a predictive accuracy throughout the whole test set trajectory. Hence, online learning was performed using a few representative input matrices comprising different time-lagged process variables; this was also helpful in understanding the relation between FIR model including the lagged variables, W and prediction accuracy. The size of the initial TS is important for learners, which should (or are preferred to) be tuned before being used in the MW frame, e.g. an initial training set is required to determine the “optimal” number of PLS components to be used during online prediction phase. It should, however, be noted that an alternative is to tune the training parameter adaptively for each query point. In the current study, both methods were used; the superscripts TS and W denote tuning using the initial TS and online, respectively. The size of the initial TS in datasets were chosen with the motivation of forming a sufficiently large training set, from which parameter tuning would be performed, and leaving aside a sufficiently large test set for obtaining a reliable estimate of test error. MW with constant W and JITL were employed to all datasets; the only exception is the application of adaptive MW (denoted by MW-D in the subscript) on DS2. FIR input matrix included process variables up to 24 lags for DS1, DS2 and DS6, while only the current sample was used for DS3, DS4 and DS5 (Table 3).

CV with 10fold and 20 repetitions 75 was employed to optimize the number of components ($L$) in PLS, and the shrinkage parameter ($\lambda$) in Lasso. When CV was employed for each MW, i.e. PLS\textsuperscript{W} and Lasso\textsuperscript{W}, 10fold CV is performed once to reduce computational burden and realize online predictions. The grid search for optimum parameters was performed on $L \in \{1, 2, \ldots \, \text{min}(25, d) \} $, in which $d$ is the total number of features in the design matrix, and $\lambda$ was varied from 10\textsuperscript{-6} to 1 logarithmically in both evaluation settings. RVM models were trained using V2 of SparseBayes package provided by Tipping, using input variables directly as the basis functions in the formulation, and all tolerance values for stopping the training algorithm were set as 10\textsuperscript{-3}\cite{63}. For offline learning scenarios, in which FIR models of different orders were considered, 10fold CV with 20 repetitions was performed to optimize the FIR model order, as well.  

\begin{table}
\centering
   \caption{Meta-parameters of the online learning scenario.}
   \label{table:3}
	\begin{tabular}{llll}
    \toprule
    Name & $n$ & Initial TS & W \\
    \midrule
	DS1  & $\{ 12, 24 \}$ & 1000 & $\{50, 100, 200, 300, 450\}$ \\
	DS2  & $\{ 0, 3, 6, 12 \}$ & 300 & $\{10, 20, \ldots \, 200\}$ \\
	DS3  & 0 & 72 & $\{10, 20, \ldots \, 70\}$ \\
	DS4  & 0 & 80 & $\{30, \ldots \, 80\}$ \\
	DS5	 & 0 & 40 & $\{10, 20, \ldots \, 40\}$ \\
	DS6/DS6p & $\{ 0, 2\}/0$ & 300 & $\{30, 40, \ldots \, 70\}$ \\
    \bottomrule
  \end{tabular} 
\end{table}

Prediction accuracy was assessed using root mean squared error (RMSE) of test observations.

\begin{align}
\mathrm{RMSE}=\sqrt{\frac{1}{N} \sum_{\mathrm{i}=1}^{N}\left(\mathrm{y}_{\mathrm{i}}-\hat{y}_{\mathrm{i}}\right)^{2}}
\label{eq:12}
\end{align}

Here, $N$ is the total number of test points. For DS6, RMSE values were further averaged over 20 repetitions and 8 different CDMs. Since RMSE depends on the unit and spread of the response variable, RMSE is inconvenient for comparing the improvement/deterioration in the prediction accuracy among different datasets. Hence, another metric, named percent increase in prediction accuracy (\%RMSE), was used to determine how the prediction accuracy of the $j^{th}$ learner may be compared to that of PLS ($\text{PLS}_{\text{MW}}^{\text{TS}}$ under online setting) under identical conditions:

\begin{align}
\% \mathrm{RMSE}_{j}=100 \times \frac{\mathrm{RMSE}_{\mathrm{PLS}}-\mathrm{RMSE}_{j}}{\mathrm{RMSE}_{\mathrm{PLS}}}
\label{eq:13}
\end{align}

Long tails in the prediction error distributions render t-test inconvenient to be used in comparing prediction performance of the learners, hence test errors of the studied learners were compared to those of $\text{PLS}_{\text{MW}}^{\text{TS}}$ using a robust t-test\cite{76} at $\alpha = 0.01$ for all datasets, expect DS5, for which $\alpha = 0.01$ was used due to the small size of its test dataset. To conduct the hypothesis tests, first, the difference between the absolute values of the test errors due to predictions of the examined learners and $\text{PLS}_{\text{MW}}^{\text{TS}}$ were computed. Then, the trimmed sample mean of the differences, consisting of $N$ observations (shown with {$\overline{X}_{t}$ in Eq. \ref{eq:14}), was subtracted from the hypothesized expected value $\mu_t = 0$ and divided by a robust standard error term ($s_w$), to determine the robust t-statistic ($T_t$). 

\begin{align}
T_{t}=\frac{(1-2 \gamma) \sqrt{N}\left(\overline{X}_{t}-\mu_{t}\right)}{s_{w}}
\label{eq:14}
\end{align}

Taking $g=\lfloor N \gamma\rfloor$ for which $\lfloor .\rfloor$ refers to flooring function, the resulting $T_t$ value is compared with $t_{\frac{\alpha}{2}, n-2, g-1}$ value to determine the significance of the hypothesis test. In the current study, a trimming proportion of $\gamma=0.1$ was used, but it was observed that using a larger value, such as $\gamma=0.2$, did not change the general picture presented here. Standard error of the trimmed samples mean was computed using “Winsorized” sample standard deviation. Winsorization refers to sorting observations in increasing order, and equating the ones with ranks smaller than or equal to $g+1$ and greater than or equal to $n-g$ with $X_{(g+1)}$ and $X_{(n-g)}$, respectively (Eq. \ref{eq:15}). The next step is to employ the conventional sample standard deviation operator on the Winsorized data to yield the Winsorized sample standard deviation (Eq. \ref{eq:16})

\begin{align}
W_{i}=\left\{\begin{array}{c}{X_{(g+1)}, \text { if } X_{i} \leq X_{(g+1)}} \\ {X_{i}, \text { if } X_{(g+1)}<X_{i}<X_{(n-g)}} \\ {X_{(n-g)}, \text { if } X_{i} \geq X_{(n-g)}}\end{array}\right.
\label{eq:15} \\
W_{i}=s_{w}^{2}=\frac{1}{n-1} \sum\left(W_{i}-\overline{W}\right)^{2} 
\label{eq:16}
\end{align}

\section{Results and Discussion}

The general analysis strategy to analyze each DS is as follows. In offline learning, RMSE values of predictions are reported with respect to the number of time-lagged measurements, followed by comparing the \%RMSE values obtained from the optimum time-lagged predictor matrices determined by CV. In online learning, RMSE and \%RMSE values obtained from fixed number of lagged measurements were reported with respect to window and neighborhood sizes for MW and JITL, respectively. In online learning part, parameter tuning was employed once for $\text{Lasso}_{\text{MW}}^{\text{TS}}$ and $\text{PLS}_{\text{MW}}^{\text{TS}}$, while CV was employed for query point for $\text{Lasso}_{\text{MW}}^{\text{W}}$ and $\text{PLS}_{\text{MW}}^{\text{W}}$.

\paragraph{Dataset 1:} In offline learning, averaging the RMSE values over 20 segments of dataset for various time lags included in the predictor matrix shows that RVM generally has the best predictive performance for both small and large TS sizes (Figs. \ref{fig:2}a and \ref{fig:2}b). While none of the learners are immune to increasing the predictor space excessively, the most significant deterioration is seen in PLS predictions. The boxplots of \%RMSE values from the models using the number of lags determined via CV for each segment show that Lasso and RVM are superior to PLS for small sized TS, but accuracy of PLS and Lasso is similar when the size of TS is enlarged (Figs. \ref{fig:2}c and d).

Under MW settings, high VIF values of the training sets justify using sparse learners (Table \ref{table:4}). RMSE and \%RMSE values obtained using 12 lagged measurements (for $n = 24$, see Fig. S2) shows that prediction accuracy of RVM exceeds that of $\text{PLS}_{\text{MW}}^{\text{TS}}$ by 5-18\% (Figs. \ref{fig:2}e and \ref{fig:2}f). $\text{Lasso}_{\text{MW}}^{\text{TS}}$ is a close runner-up, but tuning in each MW $\text{Lasso}_{\text{MW}}^{\text{W}}$ does not seem to improve its performance. Dependence of the response variable to a large number of PLS components (Table \ref{table:5}) may give an advantage to RVM and Lasso (discussed below in more detail). Accuracy of $\text{PLS}_{\text{MW}}^{\text{W}}$ exceeds that of $\text{PLS}_{\text{MW}}^{\text{TS}}$ for $\text{W} \geq 100$, but the opposite is observed for $\text{W} = 50$. Deterioration in the prediction accuracy of $\text{PLS}_{\text{MW}}^{\text{W}}$ is possibly due to high variance introduced by CV on training sets of small size, while increasing the training set size decreases the variance, and make $\text{PLS}_{\text{MW}}^{\text{W}}$ more favorable, since prediction bias is decreased via adapting to CDs. Similar results are obtained under JITL scheme (Fig. S3).

\begin{figure}[H]
\label{fig:2}
\centering
\includegraphics[scale=.8]{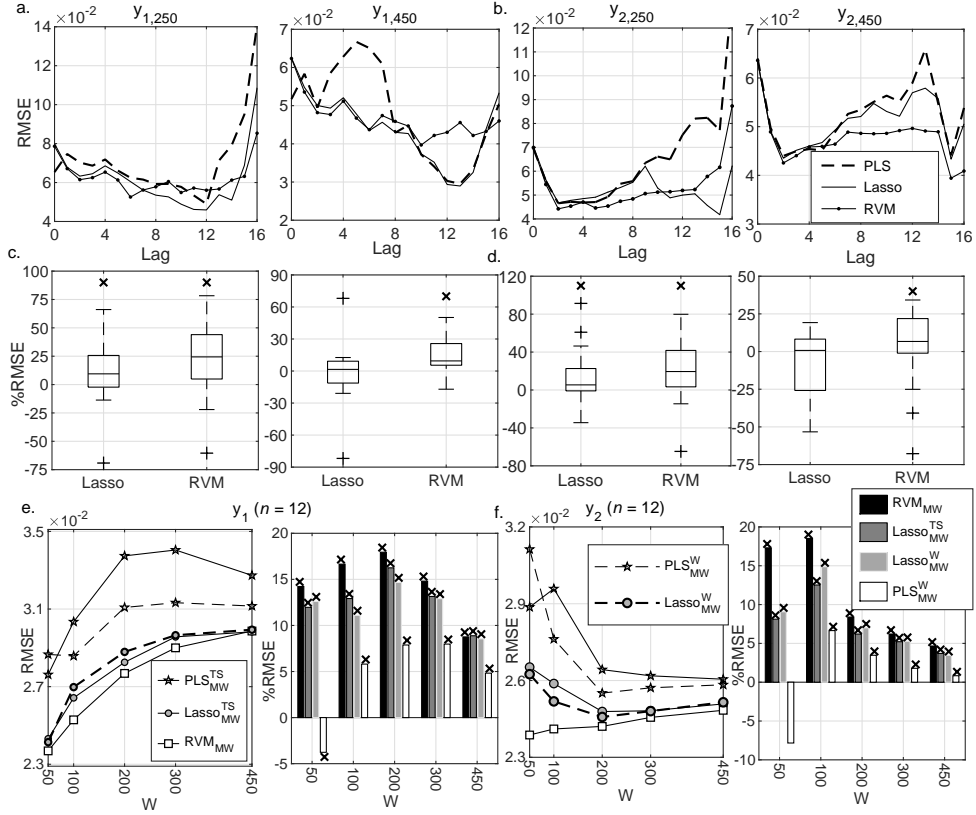}
\vskip.5\baselineskip
\caption{Results of offline (a-d) and online (e-f) learning scenarios performed on DS1. Subscripts “1” and “2” denote two response variables; subscripts “250” and “450” indicate the training set sizes. Crosses on the on the top of bars in (c-e) represent the learners with statistically significant different predictions from those of PLS.}
\end{figure}

\begin{table}[H]
   \caption{Statistics of the VIF values through the MW scheme in real process datasets.}
   \label{table:4}
   \begin{threeparttable}[b]
	\renewcommand{\TPTminimum}{\linewidth}
    \makebox[\linewidth]{%
	\begin{tabular}{lllll}
    \toprule
    Name & W & Maximum & Median & 90\textsuperscript{th} percentile \\
    \midrule
	DS1  & 50 & 1.3$\times$10\textsuperscript{4} & 314 & 1.83$\times$10\textsuperscript{3} \\
	DS2	 & 10/30/ & 7$\times$10\textsuperscript{10}/1$\times$10\textsuperscript{13}/ & 194/2.1$\times$10\textsuperscript{4}/ & 2.4$\times$10\textsuperscript{4}/7$\times$10\textsuperscript{9} \\
	     & 50/120 & 5$\times$10\textsuperscript{15}/3$\times$10\textsuperscript{6} & 2$\times$10\textsuperscript{5}/688 & 1$\times$10\textsuperscript{10}/1$\times$10\textsuperscript{5} \\
	DS3\tnote{1}  & 30 & 1$\times$10\textsuperscript{16} & 5.5 & 14.3 \\
	DS4  & 60/60/50 & 251/392/830 & 70.0/87.5/295 & 134.1/179.2/454.3 \\
	DS5	 & 40 & 15.4 & 7.2 & 10.1 \\
    \bottomrule
  \end{tabular}}    
  \begin{tablenotes}
    \item[1]{\small Infinite values were ignored.}
  \end{tablenotes}
  \end{threeparttable}
\end{table}

\begin{table}[H]
 \caption{Number of optimum latent components estimated via $\text{PLS}_{\text{MW}}^{\text{TS}}$  method in MW scheme.}
 \vskip.5\baselineskip
  \centering
    \label{table:5}
   	\begin{threeparttable}[b]
	\renewcommand{\TPTminimum}{\linewidth}
    \makebox[\linewidth]{%
  	\begin{tabular}{lllll}
    \toprule
    DS1 & \multicolumn{2}{l}{$n = 12$} & \multicolumn{2}{l}{$n = 24$}\\
    \cmidrule(r){2-5}
    & y\textsubscript{1} & y\textsubscript{2} & y\textsubscript{1} & y\textsubscript{2} \\
    & 6 & 12 & 7 & 10 \\
    DS2 & $n = 0$ & $n = 3$ & $n = 6$ & $n = 12$ \\
    \cmidrule(r){2-5}
    & 5 & 9 & 7 & 8 \\
    \midrule
    DS3 & 1 &  &  &  \\
    \midrule
    DS4 & y\textsubscript{1} & y\textsubscript{2} & y\textsubscript{3} &  \\
    & 9 & 1 & 2 & \\
    \midrule
    DS5 & y\textsubscript{1} & y\textsubscript{2} & y\textsubscript{3} &  \\
    & 4 & 4 & 1 & \\
    \midrule
    DS6\tnote{1} & $n = 3$ & $n = 0$ & \multicolumn{2}{l}{$n = 0$ (DS6p)}  \\
    & 11 & 11 & 9 & \\
    \bottomrule
  \end{tabular}}
\begin{tablenotes}
    \item[1]{\small Modes of the optimum PLS components from different CD models are shown.}
  \end{tablenotes}
  \end{threeparttable}
\end{table}

\paragraph{Dataset 2:} In offline learning scenario, prediction accuracy of PLS deteriorates faster, similar to that in DS1, as the number of predictors is increased (Fig. \ref{fig:3}a). Models obtained via CV show that RVM and Lasso predictions (except one case of RVM) are generally superior to PLS when the size of TS is small, but PLS is slightly better than RVM when training set size is increased (Fig. \ref{fig:3}b). RMSE of the test sets, predicted using PLS, Lasso and RVM at the optimum time lags obtained by CV, are equal to 0.156, 0.103 and 0.138 for N\textsubscript{TS} = 250, and equal to 0.093, 0.084, 0.100 for N\textsubscript{TS} = 400, respectively. It should be noted that these prediction errors are significantly higher than those obtained using optimum number of time lags (Fig. \ref{fig:3}a), showing that parameter tuning via random fold CV may not be efficient enough in the presence of nonstationary data\cite{65}.

In MW scheme, it is seen that prediction errors steeply increase with W if predictor matrix consists of measurements at small time lags ($\leq 3$), but prediction quality is rather insensitive to W when more lags are included (Fig. \ref{fig:3}c). This shows an interesting trade-off between the frequency of CDs and the process dynamics aimed to be modeled: process dynamics cannot be adequately modeled via including only recent observations in a FIR predictor matrix, yielding a decrease in prediction accuracy in offline learning setting. In MW models, however, exclusion of information in previous lags would make it easier to collect “homogenous” data, i.e. data from a single concept, for sufficiently small W, i.e. concentrating on recent observations would yield a more “plastic” learning frame. However, for larger W values, static relations will not be sufficient to model the behavior in a longer period of time\cite{77}. A similar behavior is not seen in JITL predictions, possibly due to training sets comprising non-contiguous data points in JITL (Fig. S4). For $n \geq 6$, RVM and Lasso improve the PLS predictions by $\sim$5-9\% and 2-7\%, respectively, while PLS and Lasso, to some extent, benefit from online parameter tuning (Figs. \ref{fig:3}c and \ref{fig:3}d). It should be noted that all the learners yielded gross errors, e.g. RMSE $\sim$0.5-1, for small $p$/W ratios, e.g. $n = 6$ and W $<$ 50 (not included within the abscissa limits in Fig. \ref{fig:3}c and \ref{fig:3}d), showing that none of the learners is immune to high collinearity problem, particularly, when sample size is small. Predictive accuracy of $\text{PLS}_{\text{MW}}^{\text{TS}}$ exceeds those of other learners for $n = 3$ with the smallest W; additionally, $\text{RVM}_{\text{MW}}$ is worse than $\text{PLS}_{\text{MW}}^{\text{TS}}$ for $n = 6$ with the smallest W, showing the superiority PLS over RVM under specific ranges of parameter values.

\begin{figure}[H]
\label{fig:3}
\centering
\includegraphics[scale=.8]{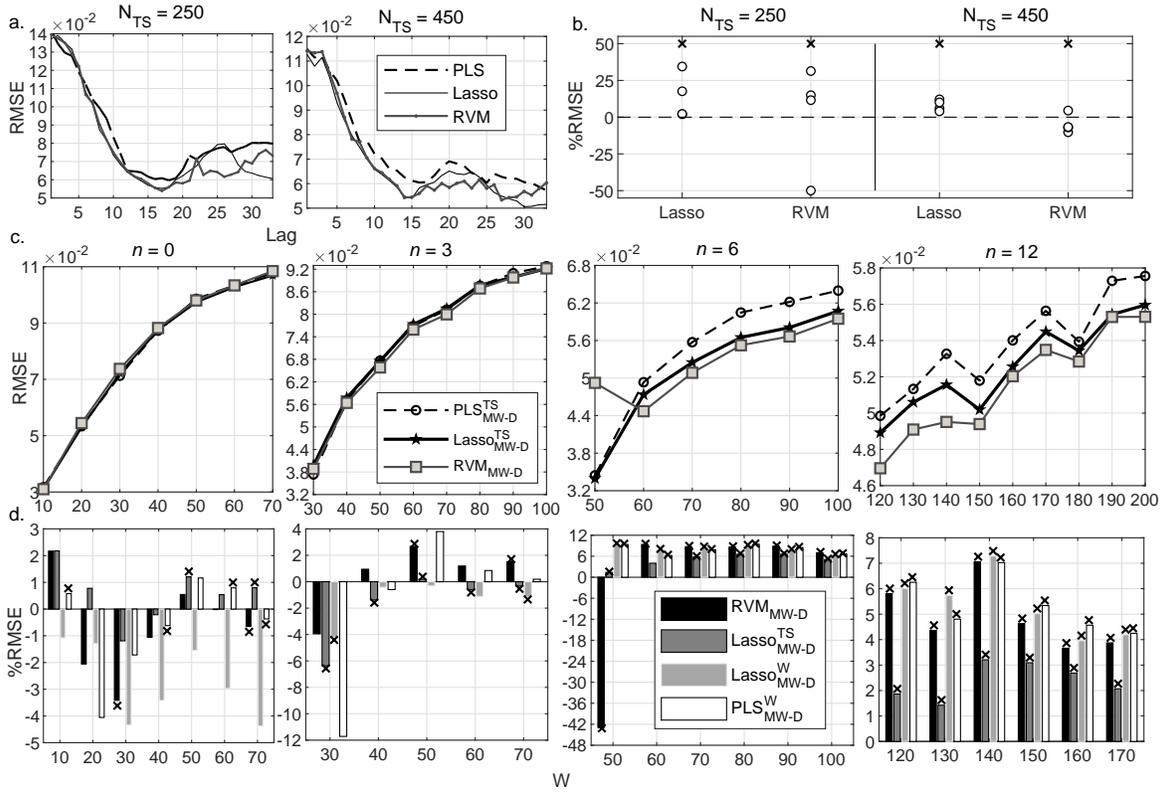}
\vskip.5\baselineskip
\caption{Results of offline (a-b) and online (c-d) learning scenarios performed on DS2. The crosses located on the top of figure in (b) and on the top of bars in (d) represent the learners with statistically significant different predictions from those of PLS. }
\end{figure}

\paragraph{Dataset 3:} While DS3 data is comprised of previous time lags up to four, including lags other than zero did not improve prediction accuracy, hence the analysis focuses on results when only the most recent measurements are considered as predictors. Offline learning test errors are the smallest for Lasso, and the largest for PLS, consistent with the results from DS1 and DS2 (Fig. \ref{fig:4}a). In both MW and JITL schemes, Lasso predictions are $\sim$3-6\% better than those of PLS, which is on par with RVM, particularly around the optimum W value (Figs. \ref{fig:4}b-c and S5). The optimum number of components in PLS models was found to be equal to unity (Table \ref{table:5}), and dependence of the quality variable to a small number of PLS components is likely to enhance the predictive performance of PLS, as suggested by Wold in the discussion to Friedman’s study\cite{29}. Similar to that in DS2, RVM yields the lowest performance at the smallest window size, and tuning PLS for every MW decreases prediction accuracy for W $<$ 40. 

\begin{figure}[H]
\label{fig:4}
\centering
\includegraphics[scale=.8]{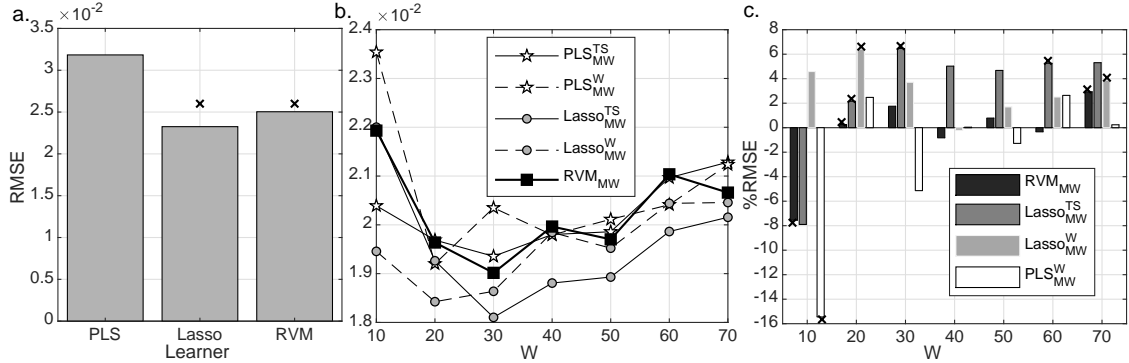}
\vskip.5\baselineskip
\caption{Results of offline (a) and online (b and c) learning scenarios performed on DS3. The crosses located on top of bars in (a) and (c) represent the learners with statistically significant different predictions from those of PLS.}
\end{figure}

\paragraph{Dataset 4:} In the offline learning scenario, all three methods are indistinguishable (Fig. \ref{fig:5}a), but reliability of prediction accuracy estimates is questionable due to the small size of the total test set ($30 \times 3 = 90$, see Table \ref{table:2})80. Similar to that in DS3, performance of PLS in MW and JITL schemes seems to depend on the number of PLS components. The number of PLS components is the highest in y\textsubscript{1}, and the lowest (equal to unity) in y\textsubscript{2} (Table \ref{table:5}); consistent with this ranking, RVM and Lasso predictions are 3-5\% and 10-16\% more accurate than those of $\text{PLS}_{\text{MW}}^{\text{TS}}$ for y\textsubscript{1} and y\textsubscript{3} for W $geq$ 40, while PLS has the best accuracy for y\textsubscript{2} (Figs. \ref{fig:5}b-d, and S6). As seen for the other datasets, $\text{PLS}_{\text{MW}}^{\text{TS}}$ seems to yield more accurate predictions, particularly, compared to those of RVM for small values of W (Figs. \ref{fig:5}c and 5d).  It is questionable that $\text{Lasso}_{\text{MW}}^{\text{W}}$ and $\text{PLS}_{\text{MW}}^{\text{W}}$ improve the prediction accuracies obtained from learners with constant tuning parameters.

\begin{figure}[H]
\label{fig:5}
\centering
\includegraphics[scale=.8]{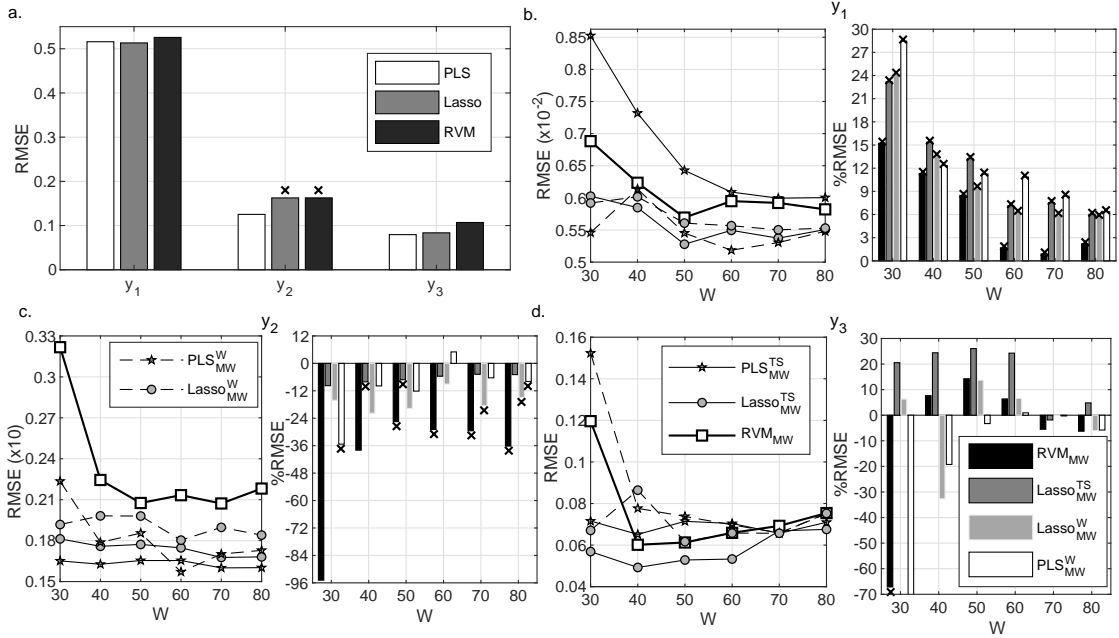}
\vskip.5\baselineskip
\caption{Results of offline (a) and online (b-d) learning scenarios performed on DS4. Offline learning results were scaled by 0.01 and 0.1 for y\textsubscript{1} and y\textsubscript{2}, respectively. \%RMSE value of y\textsubscript{3} via $\text{PLS}_{\text{MW}}^{\text{W}}$ at W = 30 is equal to -113, now shown within the y-limits in (c).}
\end{figure}

\paragraph{Dataset 5:} DS5 has the smallest maximum VIF and the smallest number of predictors among all datasets (Tables \ref{table:1} and \ref{table:2}), hence it may be expected that predictor redundancy would be low. Consistent with this expectation, all of the learners gave indistinguishable prediction performance in offline learning scenario (Fig. \ref{fig:6}a). Furthermore, optimum PLS components were found to be 4 for y\textsubscript{1} and y\textsubscript{2}, while a single optimum component was found to be sufficient for y\textsubscript{3} (Table \ref{table:5}). The optimum MW predictions were 7-12\% (based on RVM) better than those in offline learning (Figs. \ref{fig:6}b-d); this hints at the presence of mild CDs within the time frame of the test set. Differences in the performance of learners in MW setting show a strikingly similar trend to those obtained for DS2 and DS3, but with less statistical significance, mainly due to smaller size of the test set. For the smallest W values, $\text{PLS}_{\text{MW}}^{\text{TS}}$ gave the best predictions for y\textsubscript{1} and y\textsubscript{3}, but MW predictions at these W values were found to be far from optimal. RVM and Lasso were marginally superior to those of $\text{PLS}_{\text{MW}}^{\text{TS}}$ for W $\geq$ $\sim$20, while adaptive tuning deteriorated the prediction accuracy of $\text{PLS}_{\text{MW}}^{\text{W}}$, as to be expected from a process, in which harsh CDs do not take place.

\begin{figure}[H]
\label{fig:6}
\centering
\includegraphics[scale=.8]{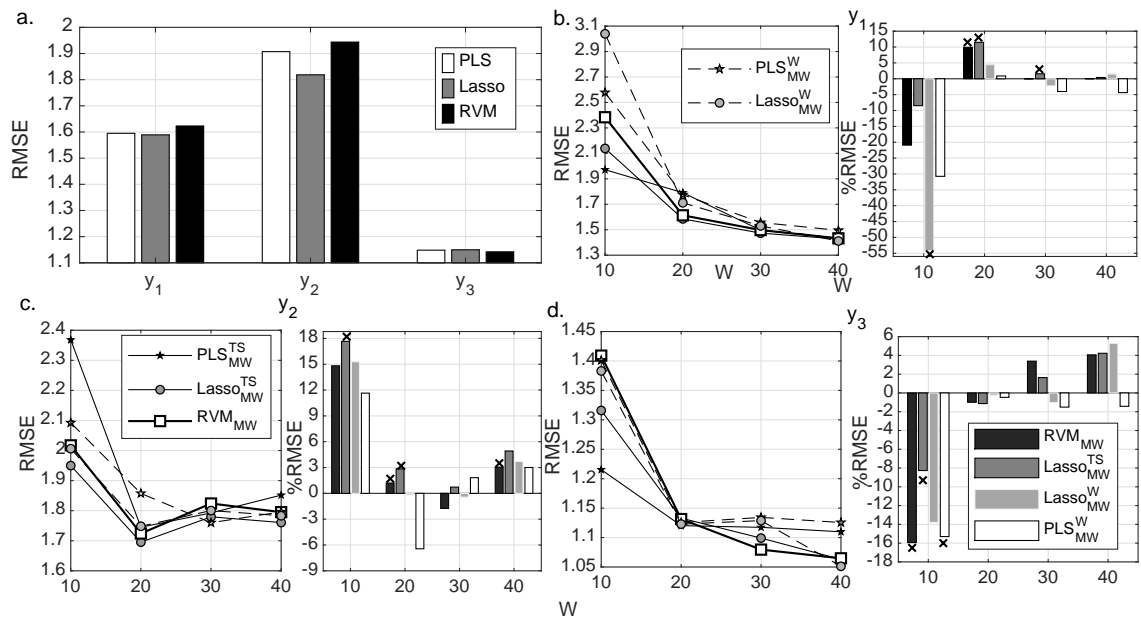}
\vskip.5\baselineskip
\caption{Results of offline (a) and online (b, c and d) learning scenarios performed on DS5.}
\end{figure}

\paragraph{Overall Analysis of the On-Line Predictions in the Experimental Datasets:} A summary of the best predictions on the industrial datasets within the range of parameters examined for MW and JITL schemes is shown in Table \ref{table:6} (see Table S6 for more details). It is to be noted that the prediction accuracies of the learning methods used in the current study are generally comparable with the representative results from the literature. MW predictions of Lasso and RVM are found to be at least as good as those PLS except one case, and 20\% trimmed mean of \%RMSE values is 4.2\% and 3.1\% for Lasso and RVM, respectively.  In JITL scheme, Lasso maintains its superiority, with a 20\% trimmed mean of \%RMSE equal to 4.8\%, but RVM does not seem to be generally advantageous compared to PLS. 

\begin{table}[H]
 \caption{Reported prediction results (RMSE values) on real datasets from the literature.}
 \vskip.5\baselineskip
  \centering
    \label{table:6}
   	\begin{threeparttable}[b]
	\renewcommand{\TPTminimum}{\linewidth}
    \makebox[\linewidth]{%
  	\begin{tabular}{lllll}
    \toprule
     & \multicolumn{2}{c}{From the literature} & \multicolumn{2}{c}{Our results}\\
    \cmidrule(r){2-3}  \cmidrule(r){4-5} 
    & Method & RMSE & MW & JITL\\ 
    \cmidrule(r){2-5}
    & & & PLS/Lasso/RVM & PLS/Lasso/RVM \\
    \midrule
    DS1	& An MLP\cite{3} & 0.030\tnote{1} & 0.027/0.024/0.024 & 0.023/0.020/0.021\\
		& 		 & 0.020 & 0.026/0.025/0.024 & 0.025/0.024/0.023 \\
	DS2 & LWKPCR\cite{17} & 0.058 & 0.0317/0.0310/0.0310 & 0.0366/0.0352/0.0394 \\
	DS3	& NUFCA\cite{79} & 0.024 & 0.0194/0.0181/0.0190 & 0.0219/0.0216/0.0219 \\
	DS4 & JLSSVR\cite{25} & - 	 & 65.9/60.2/63.5 & 65.5/60.9/62.1 \\
		& 		 & 0.849 & 1.60/1.68/2.21 & 1.62/1.57/1.84 \\
		& 		 & 0.033 & 0.065/0.053/0.060 & 0.091/0.061/0.077 \\
	DS5	& ALSSVR\cite{73} & 1.2 & 1.43/1.42/1.43 & 1.16/1.28/1.27 \\
		& 		 & 1.4 & 1.75/1.70/1.72 & 1.27/1.27/1.29 \\
		& 		 & 1.0 & 1.11/1.06/1.06 & 0.98/0.92/0.97 \\
    \bottomrule
  \end{tabular}}
\begin{tablenotes}
    \item[1]{\small For datasets with multiple response variables, each row corresponds to a different response variable and the related dataset name is aligned with the first response variable.}
  \end{tablenotes}
  \end{threeparttable}
\end{table}

Based on the analysis on DS1-5, two empirical relations between the prediction accuracies of the learners in the datasets and the learning parameters may be proposed. The first relation is between the performance of the learners and the window size in MW scheme: in most of the datasets, the relative prediction performance of PLS was found to be higher when small W values were used, while performance of PLS deteriorated at larger W. To clarify this relation, \%RMSE values from each response variable (10 response variables, in total) were averaged for W $>$ W\textsubscript{min}, in which W\textsubscript{min} corresponds to the size of the smallest MW used for each dataset, and are shown in Fig. \ref{fig:7}, while \%RMSE values for W = W\textsubscript{min} are shown in Fig. \ref{fig:7}b. It is seen that prediction accuracy in all cases except one is improved up to $\sim$15\% over $\text{PLS}_{\text{MW}}^{\text{W}}$ when $\text{RVM}_{\text{MW}}$ and $\text{Lasso}_{\text{MW}}^{\text{TS}}$ were used, while improvement due to $\text{Lasso}_{\text{MW}}^{\text{W}}$ is less pronounced. On the other hand, none of the learners seem to have a better performance than that of $\text{PLS}_{\text{MW}}^{\text{W}}$ for the smallest MW size. Consistent with the visual inspection, Wilcoxon signed-rank test on the samples yielded p-values equal to .022, .022, .11, .75 when W $>$ W\textsubscript{min} for RVM, $\text{Lasso}_{\text{MW}}^{\text{TS}}$, $\text{Lasso}_{\text{MW}}^{\text{W}}$, $\text{PLS}_{\text{MW}}^{\text{W}}$, respectively, while all four p-values were found to be greater than .30 when W = W\textsubscript{min}. This relation is less pronounced for JITL schemes (Fig. S7); the only statistically significant Wilcoxon signed-rank test ($p<.10$) was obtained for $\text{Lasso}_{\text{MW}}^{\text{TS}}$.

The second proposed relation is between the prediction accuracy and the number of estimated optimum components by $\text{PLS}_{\text{MW}}^{\text{TS}}$ in MW scheme (see Table \ref{table:5}). \%RMSE values obtained for each response variable via including the time-lagged predictors (13 cases, in total) were plotted with respect to the optimum number of PLS components for W $>$ W\textsubscript{min} (Fig. \ref{fig:7}c) and W = W\textsubscript{min} (Fig. \ref{fig:7}d). Visual inspection shows that RVM, $\text{PLS}_{\text{MW}}^{\text{W}}$, and, to a certain level, $\text{Lasso}_{\text{MW}}^{\text{W}}$ have relatively better prediction performance compared to that of $\text{PLS}_{\text{MW}}^{\text{TS}}$ as higher number of components is used in $\text{PLS}_{\text{MW}}^{\text{TS}}$. P-values for zero (Pearson/Spearman) correlation between \%RMSE and number of PLS components were found to be equal to .085/.025, .55/.57, .030/.077, .019/.042 when W $>$ W\textsubscript{min}, and .044/.030, .34/.20, .31/.14, .092/.036 when W = W\textsubscript{min} for $\text{RVM}_{\text{MW}}$, $\text{Lasso}_{\text{MW}}^{\text{TS}}$, $\text{Lasso}_{\text{MW}}^{\text{W}}$,  $\text{PLS}_{\text{MW}}^{\text{W}}$, respectively, validating the visual analysis. Lastly, representative linear fits were plotted on Figs. \ref{fig:7}c,d using MM-regression at 95\% efficiency\cite{80} due to heteroscedasticity of the data. It is seen that fitted lines for all four learners have positive slopes, consistent with the general picture presented so far. A similar relation was not, however, observed under JITL scheme.

\begin{figure}[H]
\label{fig:7}
\centering
\includegraphics[scale=.8]{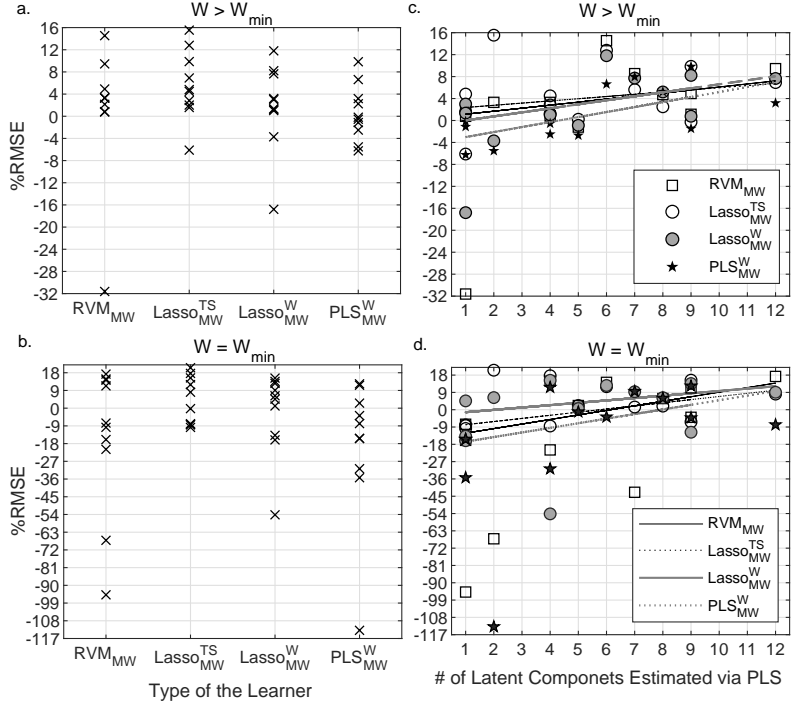}
\vskip.5\baselineskip
\caption{Summary of \%RMSE values for W $>$ W\textsubscript{min} (a) and W = W\textsubscript{min} (b) with respect to learners; \%RMSE values vs. number of optimum components for W $>$ W\textsubscript{min} (c) and W = W\textsubscript{min} (d).}
\end{figure}

\paragraph{Dataset 6:} Offline learning experiments were conducted on datasets without any real CDs, and the results are reminiscent of those obtained for DS1 and DS2: RVM has the highest prediction accuracy for all time lags, caught up by Lasso for higher time lags in offline learning scenario (Fig. \ref{fig:8}a). In MW scheme, predictions are averaged over 8 different CDMs. Including two previous time lags in the predictor matrix, RVM predictions are impressively stable for all W values, and minimum for W = 30-50, while $\text{Lasso}_{\text{MW}}^{\text{W}}$ yielded slight better predictions for W $\geq$ 60 (Fig. \ref{fig:8}b). When predictor matrix is composed of only time lag zero variables, prediction accuracies of all learners slightly deteriorated (Fig. \ref{fig:8}c), showing that FIR models with time-lagged predictors have the potential to improve prediction accuracy, also seen in DS1 and DS2.   

\begin{figure}[H]
\label{fig:8}
\centering
\includegraphics[width=\columnwidth]{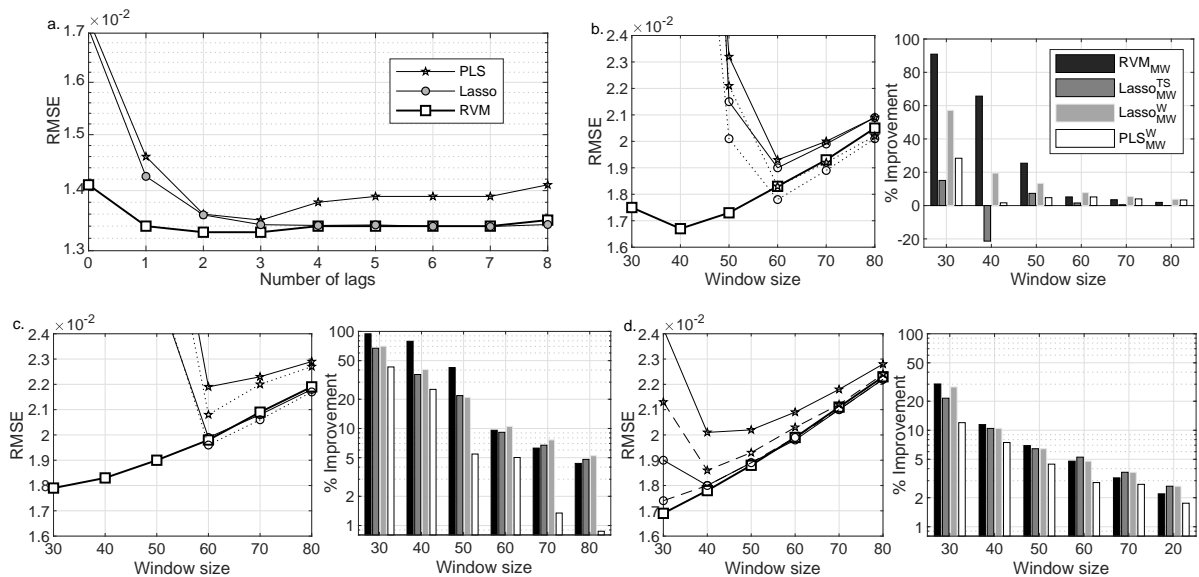}
\vskip.5\baselineskip
\caption{Results of offline learning experiments performed on simulations without CD from synthetic data (a). Results of online learning experiments performed on simulations with CD from synthetic data; predictors comprising two previous time lags (b), zero time lag (c), and a subset of process variables (d). RMSE values are averaged over all CDMs to obtain average RMSE.}
\end{figure}

Most of the real datasets are composed of preselected predictors, which are comprised of a subset of all measured (not reported) process variables. We expect that this preprocessing step should reduce the difference in the predictive performance between Lasso/RVM and PLS, in favor of PLS. To test this expectation, a further set of simulations was performed using a predictor matrix, constructed from 10 preselected process variables. Under this scenario, the optimum number of PLS components decreased from 11 to 9 (Table \ref{table:5}), and all learners, except RVM, yielded better predictions (Fig. \ref{fig:8}d). The most drastic improvement in predictions is seen in $\text{PLS}_{\text{MW}}^{\text{TS}}$; at the near optimum value of W = 40-50, RVM/Lasso predictions are $\sim$50\% better than those of $\text{PLS}_{\text{MW}}^{\text{TS}}$ for the 19 predictor case (Fig. \ref{fig:8}b), but only $\sim$10\% better for the 10 predictor case (Fig. \ref{fig:8}c). This observation is consistent with the assertion stated above: difference in predictive performances of PLS and RVM/Lasso is decreased when process variables which have relatively small importance in predicting the response variable are removed from the predictor matrix; hence, the dimension of the latent space is reduced. It should, nevertheless, be noted that predictive accuracy of RVM employed on all variables including lagged measurements (Fig. \ref{fig:8}b) still exceeds that of the preprocessed case (Fig. \ref{fig:8}d), showing that using a sparse learner in a high dimensional predictor space may yield better predictions than those would be achieved by manual selection of predictors.

\section{Conclusions}

In the current study, we evaluated PLS, Lasso and RVM as regression methods for soft sensor design in both offline and online learning schemes. Although more work is needed to increase dataset diversity, include other online learning schemes, such as temporal difference learning and recursive methods, and decipher the detailed statistical mechanisms causing the differences in prediction performances of the learners, the current study has produced a number of valuable guidelines that may be used in soft sensor design. It should be noted that in general, industrial data is preprocessed and analyzed by the practitioners to yield a small subset of the actual process variables. For that reason, not all datasets may be challenging enough in terms of collinearity to compare the learners, thus we tried to compensate for this shortcoming by including lagged measurements in design matrix.  In the offline learning schemes, RVM and Lasso were generally found to be superior to PLS. While prediction performance in online learning is superior to that in offline learning setting, a periodic application of offline learning consisting of short segments of test periods may be tolerable, particularly when using RVM on a predictor matrix including a sufficiently large number of time-lagged variables. It should, however, be noted that choosing the appropriate number of time lags using CV requires further investigation. In online learning schemes, predictions of PLS were observed to be superior to those of RVM and Lasso for cases when small training sets and few PLS components were used. On the other hand, the optimum moving window sizes and number of PLS components in the current datasets were usually found to be slightly higher than the range for which PLS gave the best predictions. Overall, the highest online prediction accuracy was achieved usually by Lasso and RVM, exceeding that of PLS by $\sim$5\%, while online tuning did not seem to offer substantial advantage over off-line tuned learners. Occasional unstable predictions of RVM for small training sets is to be noted; in the absence of a CV step for parameter tuning, RVM may be more prone to overfit. A potential source of bias in the industrial datasets may be due to using small subsets of process variables, possibly increasing the relative performance of PLS. As shown via simulations, using sparse learners on a larger set of predictors increased the prediction accuracy more than that of PLS. While RVM was found to yield better predictions than Lasso under this scenario, more experiments on industrial data including larger number of process variables are required before generalizing. Prediction accuracies of learners showed more significant differences in the MW scheme compared to JITL; this point requires further investigation. Based on the current findings, we recommend using Lasso, or RVM, only if a sufficiently large dataset is available to validate its performance, in designing a soft sensor.

\bibliographystyle{abbrv}
\bibliography{ms} 
\end{document}